\documentclass[conference]{IEEEtran}
\IEEEoverridecommandlockouts

\usepackage{amsmath,amssymb,amsfonts}
\usepackage{algorithmic}
\usepackage{graphicx}
\usepackage{newtxtext}
\usepackage{textcomp}
\usepackage{threeparttable}
\usepackage[caption=false, font=footnotesize]{subfig}
\def\BibTeX{{\rm B\kern-.05em{\sc i\kern-.025em b}\kern-.08em
    T\kern-.1667em\lower.7ex\hbox{E}\kern-.125emX}}
\usepackage[noadjust]{cite}
\usepackage[hang,flushmargin]{footmisc}
\usepackage{hyperref}
\hypersetup{
      breaklinks=true,  
      colorlinks=true,
      urlcolor=blue,
      linkcolor=blue,
      citecolor=blue,
}

\makeatletter
\def\footnoterule{\kern-3\p@
  \hrule \@width 3.5in \kern 2.6\p@} 
\makeatother
\begin{document}

\title{\fontsize{12pt}{12pt}\selectfont \textbf{DEEP LEARNING WITH CONVOLUTIONAL NEURAL NETWORKS FOR DECODING AND VISUALIZATION OF EEG PATHOLOGY}
\thanks{The present work was partly funded by the cluster of excellence BrainLinks-BrainTools (DFG grant EXC 1086) to the University of Freiburg.}
}

\DeclareRobustCommand*{\IEEEauthorrefmark}[1]{%
  \raisebox{0pt}[0pt][0pt]{\textsuperscript{\footnotesize #1}}%
}

\author{
    \IEEEauthorblockN{\fontsize{12pt}{12pt}\selectfont
      \textit{
        R. Schirrmeister\IEEEauthorrefmark{1},
        L. Gemein\IEEEauthorrefmark{2},
        K. Eggensperger\IEEEauthorrefmark{3},
        F. Hutter\IEEEauthorrefmark{3} and
        T. Ball\IEEEauthorrefmark{1}} \ \\ \ \\}
    \IEEEauthorblockN{\fontsize{10pt}{10pt}\selectfont
        \IEEEauthorrefmark{1} Translational Neurotechnology Lab, Medical Center --- University of Freiburg, Germany \\
        \IEEEauthorrefmark{2} Department of Computer Science, University of Freiburg, Germany\\
        \IEEEauthorrefmark{3} Machine Learning Lab, University of Freiburg, Germany \ \\ \ \\}
    \IEEEauthorblockN{\fontsize{10pt}{10pt}\selectfont
        \{robin.schirrmeister, tonio.ball\}@uniklinik-freiburg.de,  \{gemeinl, eggenspk, fh\}@cs.uni-freiburg.de\\
        }
}
\maketitle


\begin{abstract}
We apply convolutional neural networks (ConvNets) to the task of distinguishing pathological from normal EEG recordings in the Temple University Hospital EEG Abnormal Corpus.
We use  two basic, shallow and deep ConvNet architectures recently shown to decode task-related information from EEG at least as well as established algorithms designed for this purpose.
In decoding EEG pathology, both ConvNets reached substantially better accuracies (about 6\% better, $\approx$85\% vs. $\approx$79\%) than the only published result for this dataset, and were still better when using only 1 minute of each recording for training and only six seconds of each recording for testing.
We used automated methods to optimize architectural hyperparameters and found intriguingly different ConvNet architectures, e.g., with max pooling as the only nonlinearity.
Visualizations of the ConvNet decoding behavior showed that they used spectral power changes in the delta (0-4 Hz) and theta (4-8 Hz) frequency range, possibly alongside other features, consistent with expectations derived from spectral analysis of the EEG data and from the textual medical reports.
Analysis of the textual medical reports also highlighted the potential for accuracy increases by integrating contextual information, such as the age of subjects.
In summary, the ConvNets and visualization techniques used in this study constitute a next step towards clinically useful automated EEG diagnosis and establish a new baseline for future work on this topic.
\end{abstract}

\section{Introduction}

Electroencephalography (EEG) is widely used in clinical practice because of its low cost and its lack of side effects due to its noninvasive nature.
It is important both as a screening method as well as for hypothesis-based diagnostics, e.g., in epilepsy or stroke.
One of the main limitations of using EEG for diagnostics is the required time and specialized knowledge of experts that need to be well-trained on EEG diagnostics to reach reliable results.
Therefore, a machine-learning approach that aids in the diagnostic process could make EEG diagnosis more widely accessible, reduce time and effort for clinicians and potentially make diagnoses more accurate.

In recent years researchers have increasingly addressed the field of computer-aided EEG diagnosis.
So far, the applications were mostly limited to specific diagnoses such as Alzheimer's disease \cite{lehmann_application_2007}, depression \cite{cai_pervasive_2016,hosseinifard_classifying_2013}, traumatic brain injuries \cite{albert_automatic_2016}, or stroke \cite{giri_ischemic_2016}.
They used a large variety of machine-learning techniques, including k-nearest neighbors, random forests, support vector machines, linear discriminant analysis, logistic regression, neural networks, and more. This large variety of used methods indicates that the search for the best decoding approach for diverse types of EEG diagnosis is still ongoing.

To overcome the lack of large datasets representative of the variety of EEG-diagnosable diseases and the heterogeneity of clinical populations, the Temple University Hospital (TUH) has published an unprecedented public dataset of clinical EEG recordings \cite{obeid_temple_2016}.
From this dataset with over 16000  clinical recordings, the TUH Abnormal EEG Corpus with about 3000 recordings has been created specifically to foster the development of methods for distinguishing pathological from normal EEG.
Due to its size and rich annotation, this data set has a lot of potential to contribute to the progress of automated EEG diagnosis.
Baseline results on this dataset have already been reported by TUH using a convolutional neural network (ConvNet) with multiple fully connected layers that uses precomputed EEG bandpower-based features as input and reached 78.8\% accuracy \cite{lopez_automated_2017}.

Deep learning approaches recently receive increasing attention in many types of machine learning problems in healthcare \cite{miotto_deep_2017}.
Deep ConvNets trained end-to end from the raw signals are a promising deep learning technique.
These ConvNets exploit the hierarchical structure present in many natural signals.
Recently, deep ConvNets trained end-to-end were, for example, able to more accurately diagnose skin cancer types from images than human dermatologists \cite{esteva_dermatologist-level_2017} and could segment retinal vessels better than human annotators \cite{maninis_deep_2016}.

Deep ConvNets are nowadays also being applied to EEG analyses, such as decoding task-related information from EEG \cite{schirrmeister_deep_2017, hajinoroozi_eeg-based_2016, lawhern_eegnet:_2016, manor_convolutional_2015, stober_deep_2015, tabar_novel_2017}.
We have recently developed and validated the Braindecode toolbox\footnote{\url{https://github.com/robintibor/braindecode}, code to reproduce the results of this study is available under \url{https://github.com/robintibor/auto-eeg-diagnosis-example}} for this purpose, and showed that the performance of deep ConvNets trained end-to-end is comparable to that of algorithms using hand-engineered features to decode task-related information.
We also introduced novel visualization methods to gain a better understanding of ConvNet decoding behavior.
 
 In this study, we apply deep ConvNets to the problem of distinguishing normal from pathological EEG on the TUH EEG Abnormal Corpus and show that they can reach better accuracies than the only published baseline result we are aware of, establishing a new improved baseline for future work in this field.
 
\section{Methods}
\subsection{EEG ConvNet architectures and training}
\begin{figure*}[htbp]
\centering
\includegraphics[width=\linewidth]{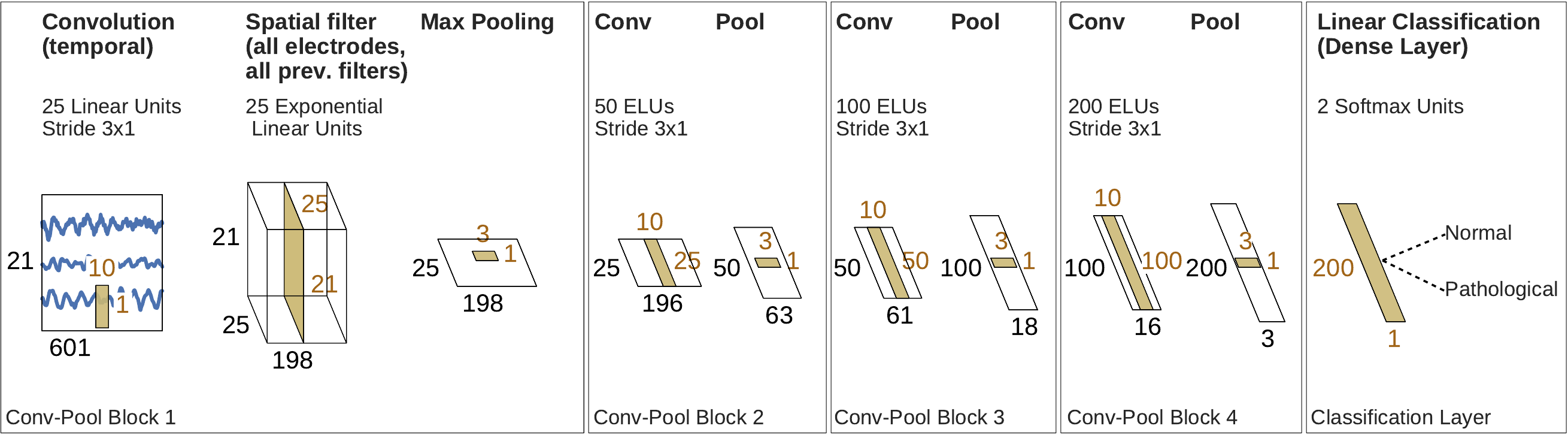}
\caption{\textbf{Deep ConvNet architecture.}
Black cuboids: inputs/feature maps; brown cuboids: convolution/pooling kernels.
The corresponding sizes are indicated in black and brown, respectively.
Each spatial filter has weights for all possible pairs of electrodes with filters of the preceding temporal convolution.
Note that in these schematics, proportions of maps and kernels are only approximate.
\label{fig:deep-convnet}}
\end{figure*}

\begin{figure}[htbp]
\centering
\includegraphics[width=\linewidth]{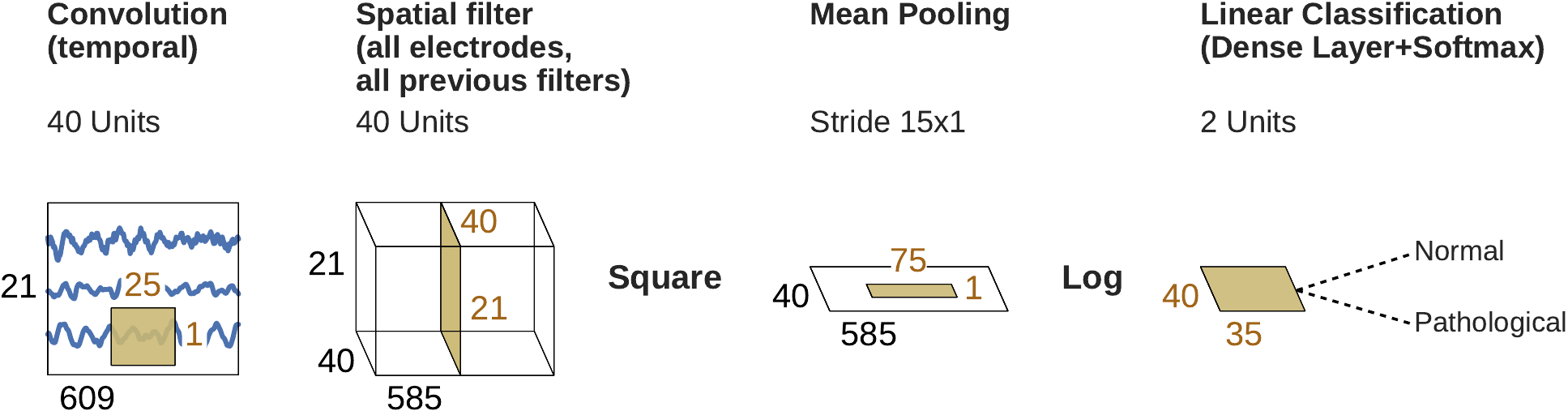}
\caption{\textbf{Shallow ConvNet architecture.} 
Conventions as in Fig. \ref{fig:deep-convnet}
\label{fig:shallow-convnet}}
\end{figure}

We used two convolutional network architectures, for both of which we recently showed that they decode task-related information from raw time-domain EEG with at least as good accuracies as previous state-of-the-art algorithms relying on hand-engineered features \cite{schirrmeister_deep_2017}.
Our deep ConvNet is a fairly generic architecture (Fig. \ref{fig:deep-convnet}), while our shallow ConvNet is specifically tailored to decode band-power features (Fig. \ref{fig:shallow-convnet}).
For more details on these models, see \cite{schirrmeister_deep_2017}.
To accommodate the longer duration of the EEG inputs as compared to our previous study, we adapted the architectures by changing the final layer filter length so the ConvNets have an input length of about 600 input samples, which correspond to 6 seconds for the 100 Hz EEG input.
Additionally, we moved the pooling strides of the deep ConvNet to the convolutional layers directly before each pooling. This modification, which we initially considered a mistake, allowed us to grow the ConvNet input length without strongly increased computation times and provided good accuracies in preliminary experiments on the training data; therefore we decided to keep it.
We optimized the ConvNet parameters using stochastic gradient descent with the optimizer Adam \cite{kingma_adam:_2015}.
To make best use of the available data, we trained the ConvNets on maximally overlapping time crops using cropped training as described by \cite{schirrmeister_deep_2017}. Code to reproduce the results of this study is available under \url{https://github.com/robintibor/auto-eeg-diagnosis-example}.

\subsection{Decoding from reduced EEG time segments}
We also evaluated the ConvNets on reduced versions of the datasets, using only the first 1, 2, 4, 8, or 16 minutes after the first minute of the recording (the first minute of the recordings was always excluded because it appeared to be more prone to artifact contamination than the later time windows).
We reduced either only the training data, only the test data, or both.
These analyses were carried out to study how long EEG recordings need to be for training and for predicting EEG pathologies with good accuracies.

\subsection{Automatic architecture optimization}
We also carried out a preliminary study of automatic architecture optimization to further improve our ConvNet architectures.
To that end, we used the automatic hyperparameter optimization algorithm SMAC \cite{hutter_sequential_2011} to optimize architecture hyperparameters of the deep and shallow ConvNets, such as filter lengths, strides and types of nonlinearities.
As the objective function to optimize via SMAC, we used 10-fold cross-validation performance obtained on the first 1500 recordings of the training data (using each fold as an instance for SMAC to speed up the optimization).
We set a time limit of 3.5 hours for each configuration run on a single fold.
Runs that timed out or crashed (e.g., networks configurations that did not fit in GPU memory) were scored with an accuracy of 0\%.

\subsection{Visualizations of the spectral differences between normal and pathological recordings}
To understand class-specific spectral characteristics in the EEG recordings, we analyzed band powers in five frequency ranges: delta (0--4 Hz), theta (4--8 Hz), alpha (8--14 Hz), low beta (14--20 Hz), high beta (20--30 Hz) and low gamma (30--50 Hz).

For this, we performed the following steps:
\begin{enumerate}
 \item Compute a short-term Fourier transformation with window size 12 seconds and overlap 6 seconds using a Blackman-Harris window.
 \item Compute the median over all band powers of all windows and recordings in each frequency bin; independently for pathological and normal recordings.
 \item Compute the log ratio of these median band powers of the pathological and normal recordings.
 \item Compute the mean log ratio over all frequency bins in each desired frequency range for each electrode.
 \item Visualize the resulting log ratios as a topographical map.
\end{enumerate}

\subsection{Visualizations based on the effects of amplitude perturbations on decoding decisions}
Understanding the ConvNet behavior and decoding predictions is important for automatic EEG diagnosis to become practically useful as an assistive diagnosis technology.
To better understand the ConvNets used in this study, we used the input-perturbation network-prediction correlation maps that we recently developed specifically for ConvNets for EEG decoding.
This method shows the effect of perturbing the input amplitudes in different frequencies on the ConvNet decoding predictions.
This visualization can provide spatial maps that show where on the scalp an amplitude change in a given frequency range correlates negatively or positively with the ConvNet classification decision.
For more details, see \cite{schirrmeister_deep_2017}.

\subsection{Analysis of word frequencies in the medical reports}
Furthermore, to better understand what kind of recordings are easier or harder for the ConvNets to correctly decode, we analyzed the textual clinical reports of each recording as included in the TUH Abnormal EEG Corpus. 
Specifically, we investigated which words were relatively more or less frequent in the incorrectly compared with the correctly predicted recordings.
We performed this analysis independently for both the normal and the pathological class of recordings.
Concretely, for each class, we first computed the relative frequencies $f_{i-}$ for each word $w_{i-}$ in the incorrectly predicted recordings, i.e.:
$f_{i-} = \frac{|w_{i-}|}{\sum_{i}|w_{i-}|}$, where $|w_{i-}|$ denotes the number of occurrences for word $w_i$ in the incorrectly predicted recordings. 
We then computed the frequencies $f_{i+}$ in the same way and computed the ratios $r_i=f_{i-}/f_{i+}$.
Finally, we analyzed words with very large ratios ($\gg1$) and very small ratios ($\ll1$) by inspecting the contexts of their occurrences in the  clinical reports.
This allowed us to gain insights into which clinical/contextual aspects of the recordings correlated with ConvNets failures.

\subsection{Dataset}

\begin{table}
\begin{threeparttable}[b]
\caption{TUH EEG Abnormal Corpus 1.1.2 Statistics\tnote{1}}
\begin{center}
\begin{tabular}{|l|l|c|c|}
\hline
 & & \textbf{Files} & \textbf{Patients} \\
 \hline
  \textbf{Train} & Normal& 1379 (50\%) & 1238 (58\%) \\
& Pathological & 1361(50\%) & 894 (42\%) \\
 & Rater Agreement\tnote{2} & 2704 (99\%) & 2107 (97\%)\\
 & Rater Disagreement \tnote{2} & 36 (1\%) & 25 (0\%) \\
 \hline
\textbf{Evaluation}& Normal & 150 (54\%) & 148 (58\%)\\
 & Pathological & 127 (46\%) &  105 (42\%) \\
 & Rater Agreement \tnote{2} & 277 (100\%) & 253 (100\%) \\
 & Rater Disagreement \tnote{2} & 0 (0\%) & 0 (0\%) \\

\hline
\end{tabular}

 \begin{tablenotes}
\item [1] Obtained from \url{https://www.isip.piconepress.com/projects/tuh_eeg/}.
\item [2] These fields refer to the agreement between the annotator of the file and the medical report written by a certified neurologist.
\end{tablenotes}
\end{center}
\label{tab:dataset}
\end{threeparttable}
\end{table}

The Temple University Hospital (TUH) EEG Abnormal Corpus 1.1.2 is a dataset of manually labeled normal and pathological clinical EEG recordings.
It is taken from the TUH EEG Data Corpus which contains over 16000 clinical recordings of more than 10000 subjects from over 12 years \cite{obeid_temple_2016}.
The Abnormal Corpus contains 3017 recordings, 1529 of which were labeled normal and 1488 of which were labeled pathological.
The Corpus was split into a training and evaluation set, see Table \ref{tab:dataset}.

Recordings were acquired from  at least 21 standard electrode positions and with a sampling rate of in most cases 250 Hz.
Per recording, there are around 20 minutes of EEG data.
The inter-rater agreement on between the medical report of a certified neurologist and another annotator was 99\% for the training recordings and 100\% for the evaluation recordings.

\subsection{Preprocessing}
We minimally preprocessed the data with these steps:
\begin{enumerate}
 \item Select a subset of 21 electrodes present in all recordings.
 \item Remove the first minute of each recording as it contained stronger artifacts.
 \item Use only up to 20 minutes of the remaining recording to speed up the computations.
 \item Clip the amplitude values to the range of $\pm800$ $\mu V$ to reduce the effects of strong artifacts.
 \item Resample the data to 100 Hz to further speed up the computation. 
\end{enumerate}

\section{Results}
\subsection{Deep and shallow ConvNets reached state-of-the-art results}

\begin{table}[htbp]
\caption{Decoding accuracies for discriminating normal and pathological EEG with deep and shallow ConvNets.}
\begin{center}
\begin{tabular}{|c|c|c|c|c|}
 \hline
 & Accuracy & Sensitivity & Specificity  & Crop-accuracy \\
\hline
Baseline & 78.8 &  75.4 & 81.9 & n.a. \\
Deep & 85.4 & 75.1 & 94.1 & 82.5 \\
Shallow & 84.5 & 77.3 & 90.5 & 81.7 \\
Linear  & 51.4 & 20.9 & 77.3 & 50.2 \\
\hline
\end{tabular}
\label{tab:main-results}
\end{center}

\begin{tablenotes}
\small
\item Results on the evaluation set of the TUH EEG Abnormal Corpus.
For deep and shallow ConvNets, mean over five independent runs with different random seeds.
Sensitivity and specificity are, as commonly defined, the ratio of the number of true positives to the number of all positives and the ration of the number of true negatives to the number of all negatives, respectively.
Deep and shallow ConvNet outperformed the feature-based deep learning baseline  \cite{lopez_automated_2017}.
n.a.: not applicable.
\end{tablenotes}
\end{table}
Both the deep and the shallow ConvNet outperformed the only results published on the TUH Abnormal EEG Corpus so far (see Table \ref{tab:main-results}).
Both ConvNets were more than 5\% better than the baseline method of a convolutional network that included multiple fully connected layers at the end and took precomputed EEG features of an entire recording as one input \cite{lopez_automated_2017} \footnote{Note that the baseline was evaluated on an older version of the Corpus that has since been corrected to not contain the same patient in training and test recordings among other things.}.
The ConvNets as applied here reduced the error rate from about 21\% to about 15\%. We also tested a linear classifier on the same 6-second inputs as our ConvNets. The linear classifier did not  reach accuracies substantially different from chance (51.4\%).

\begin{figure}[htbp]
\centering
\subfloat[]{%
\includegraphics[width=0.408\linewidth]{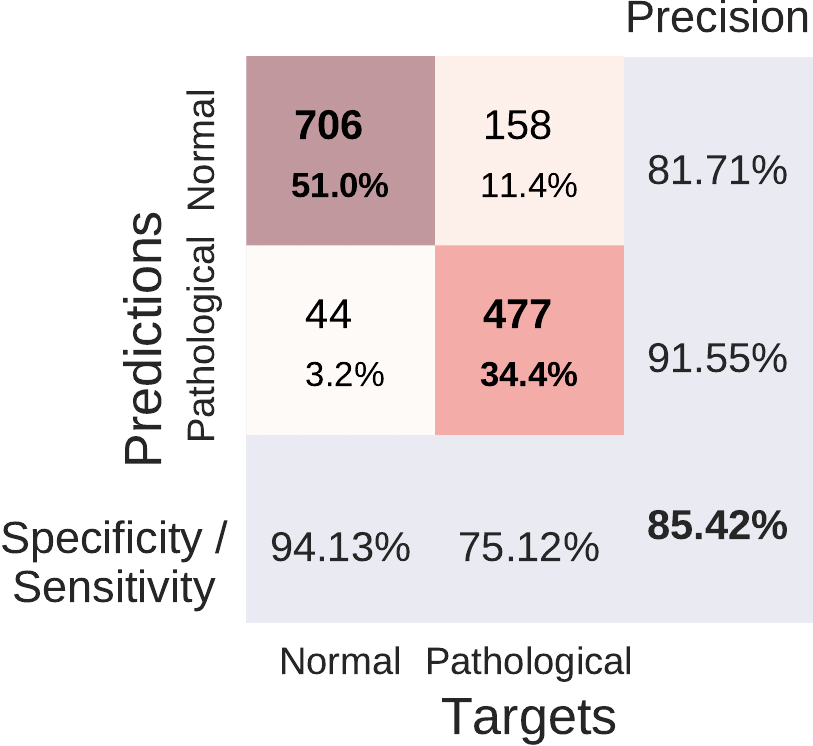}
}\hfill
\subfloat[]{%
\includegraphics[width=0.562\linewidth]{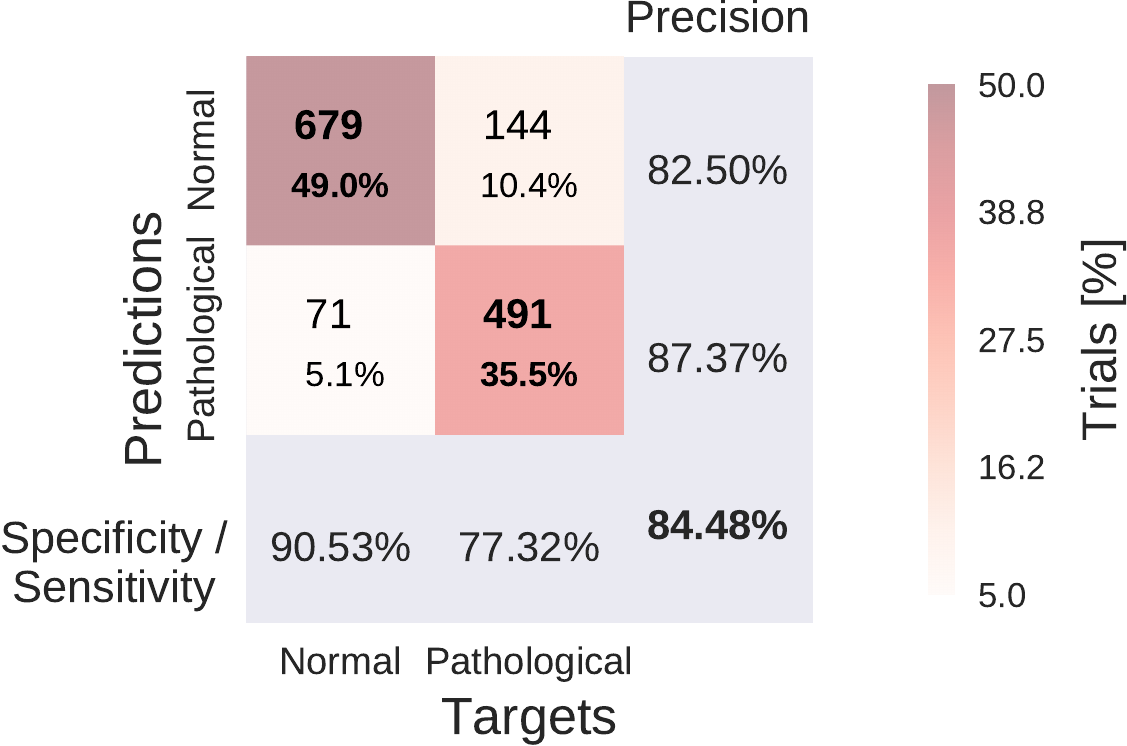}
}
\caption{\textbf{Confusion Matrices for deep and shallow ConvNets}, summed over five independent runs.
Each entry of row r and column c for upper-left 2x2-square: Number of trials of target r predicted as class c (also written in percent of all trials).
Bold diagonal corresponds to correctly predicted trials for both classes. Percentages and colors indicate fraction of trials in each cell relative to all trials.
The lower-right value: overall accuracy. The first two values in the bottom row correspond to sensitivity and specificity.
Rightmost column corresponds to precision defined as the number of trials correctly predicted for class r/number of trials predicted as class r.
.}
\label{fig:confmats}
\end{figure}

Both of our ConvNets made more errors on the pathological recordings, as can be seen from Fig. \ref{fig:confmats}.
Both ConvNets reached a specificity of above 90\% and a sensitivity of about 75-78\%.
Confusion matrices between both approaches were very similar.
Relative to the baseline, they reached a similar sensitivity (0.3\% smaller for the deep ConvNet, 1.9\% higher for the shallow ConvNet), and a higher specificity (12.2\% higher for the deep ConvNet and 8.6\% higher for the shallow ConvNet).

Interestingly, both of our ConvNet architectures already reached higher accuracies than the baseline when evaluating single predictions from 6-second crops.
The average per-crop accuracy of individual predictions was only about 3\% lower than average per-recording accuracy (averaged predictions of all crops in a recording).
Furthermore, the individual prediction accuracies were already about 3\% higher than the per-recording accuracies of the baseline.
This implies that predictions with high accuracies can be made from just 6 seconds of EEG data.

\subsection{Deep ConvNet reached best accuracies using only 1 minute per test-recording}
\begin{figure}[htbp]
\centerline{\includegraphics[width=\linewidth]{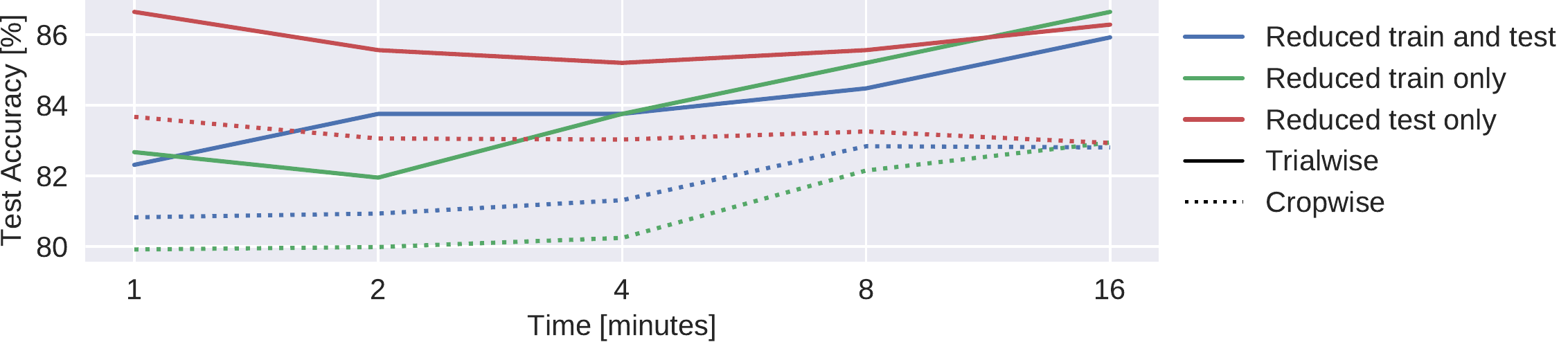}}
\caption{\textbf{Results on reduced datasets for deep ConvNet.} Train and/or test (evaluation) dataset was reduced from 20 minutes per recording to 1,2,4,8, or 16 minutes per recording, results are shown on the test set. Notably, when only reducing the duration of the test set recordings, maximal accuracies were observed  when using just 1 minute. We note that these results are each based on one run only; the slightly better performance than in Table \ref{tab:main-results} may thus be due to noise.}
\label{fig:reduced-time}
\end{figure}

\begin{figure}[htbp]
\centerline{\includegraphics[width=\linewidth]{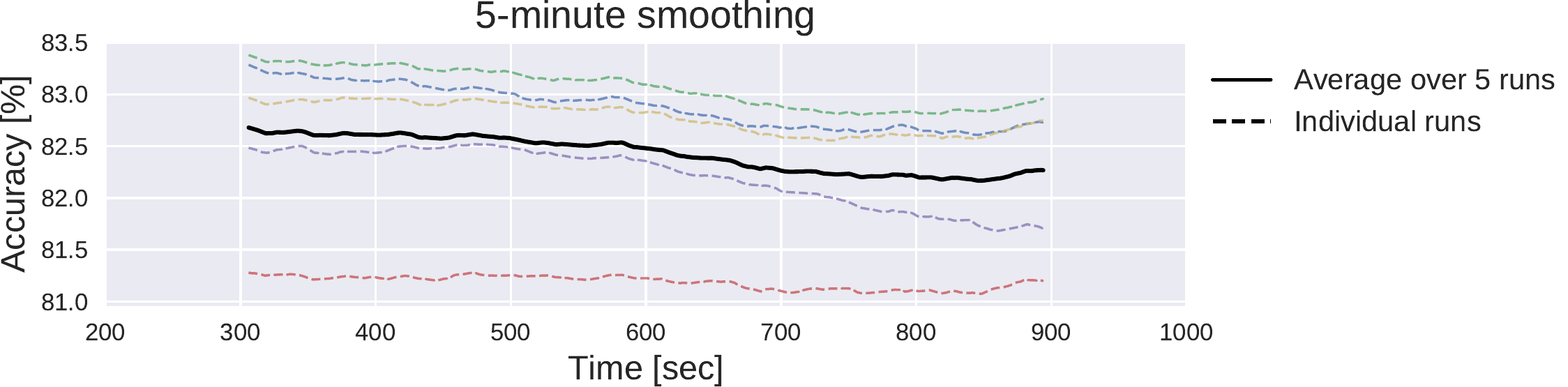}}
\caption{\textbf{Moving average of cropwise accuracies for the deep ConvNet.}
5-minute moving averages of the cropwise accuracies of the deep ConvNet, averaged over all test set recordings.
Dashed lines represent 5 individual training runs with different random seeds, solid black line represents mean over results for these runs.
x-axis shows center of 5-minute averaging window.}
\label{fig:pred-time-course}
\end{figure}

Deep ConvNets already reached their best trialwise accuracies with only one minute of data used for the prediction. While the reduction of the amount of length  of the training data led to crop- and trialwise accuracy decreases on the test data, reductions in the test data did not have such an effect (see Fig. \ref{fig:reduced-time}).
Remarkably, both crop- and trialwise accuracies slightly decreased when going from 1 minute to 2 or 4 minutes of test data.
To investigate whether earlier parts of the recordings might be more informative, we also computed a 5-minute moving average of the cropwise accuracies on the test data for the Deep ConvNet trained on the full data.
We show the average over all recordings for these moving averages in Fig. \ref{fig:pred-time-course}.
Noticeably, as expected, accuracies slightly decreased with increasing recording time.
However, the decrease is below 0.5\% and thus should be interpreted cautiously.

\subsection{Architecture optimization yielded unexpected new models}

\begin{figure}[htbp]
\centering
\includegraphics[width=\linewidth]{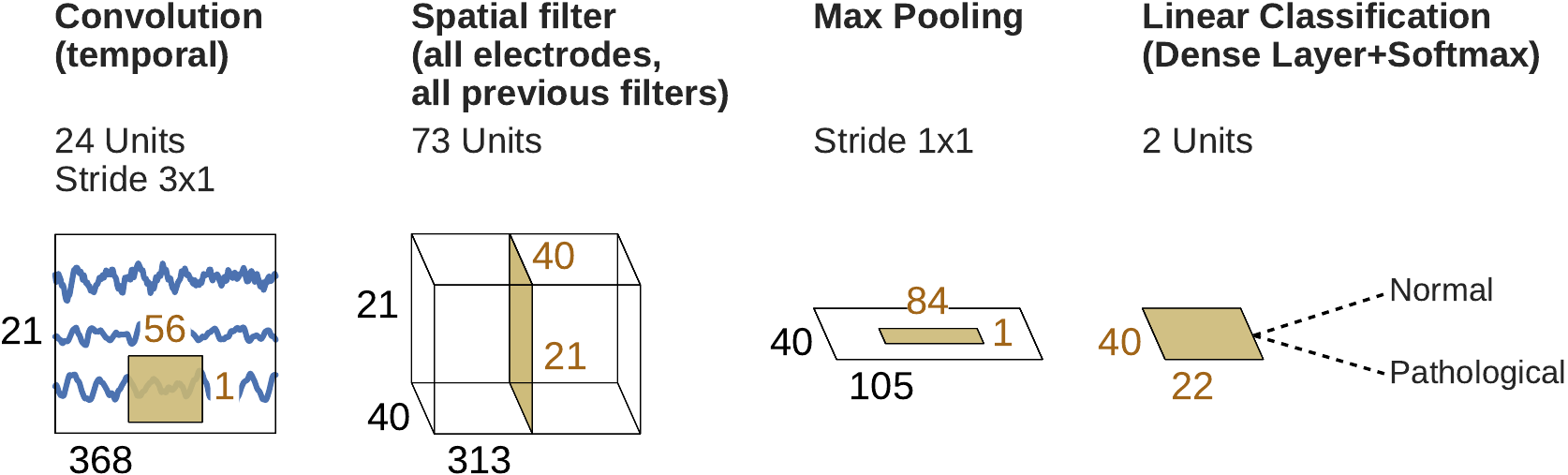}
\caption{\textbf{Final shallow ConvNet architecture selected by SMAC.}
Conventions as in Fig. \ref{fig:shallow-convnet}.
Note that max pooling is the only nonlinearity SMAC decided to use.}
\label{fig:smac-convnet}
\end{figure}

\begin{table}
\begin{center}
\begin{threeparttable}[b]
\caption{Decoding accuracies on training and test set}
\begin{tabular}{|c|c|c|c|c|c|}
 \hline
 \textbf{}& & \multicolumn{2}{|c|}{\textbf{Train\tnote{1}}}& \multicolumn{2}{|c|}{\textbf{Test}} \\
 \cline{3-6}
 & \textbf{Config} & \textbf{Trial} & \textbf{Crop} & \textbf{Trial} & \textbf{Crop} \\
\hline
\textbf{Deep} & Default & 84.2 &  81.6 & 85.4 & 82.5\\
 & Optimized & 86.3 & 80.9 & 84.5 & 81.3 \\
 \hline
\textbf{Shallow}  & Default & 84.5 &  82.1 & 84.5 & 81.7 \\
& Optimized & 85.9 & 80.3 & 83.0 & 79.8 \\
\hline
\end{tabular}
\label{tab:smac-results}
 \begin{tablenotes}
\item [1] 10-fold cross-validation on the 1500 chronologically earliest recordings of the training data
\end{tablenotes}
\end{threeparttable}
\end{center}
\end{table}

The models discovered by automated architecture optimization were markedly different from our original deep and shallow ConvNets, which were designed based on the experience in a previous study on decoding of task-related information from EEG \cite{schirrmeister_deep_2017}.
For example, the optimized architectures used only 1.8 and 3.7 seconds of EEG data for the optimized deep and shallow ConvNet, respectively, in contrast to about 6 seconds in the original versions.
While the improved performance of these modified architectures for the 10-fold cross-validation on the training dataset (2.1\% and 1.4\% improvement for deep and shallow ConvNets, respectively) did not generalize to the evaluation set (0.9\% and  1.5\% deterioration for deep and shallow ConvNets, respectively, see Table \ref{tab:smac-results}), the modifications to the original network architectures already provided interesting insights for further exploration:
For example, in the case of the shallow ConvNet, the modified architecture did not use any of the original nonlinearities, but used max pooling as the only nonlinearity (see Fig. \ref{fig:smac-convnet}), a configuration we had not considered in our  manual search so far. 

\subsection{Power spectra and ConvNet visualizations}

\begin{figure*}[htbp]
\centering
\subfloat[Pathological vs. normal relative spectral bandpower differences for the training set. Shown is the logarithm of the ratio of the median bandpower of the pathological  vs. normal (according to the experts' ratings) EEG recordings.]{%
\includegraphics[width=0.8\textwidth]{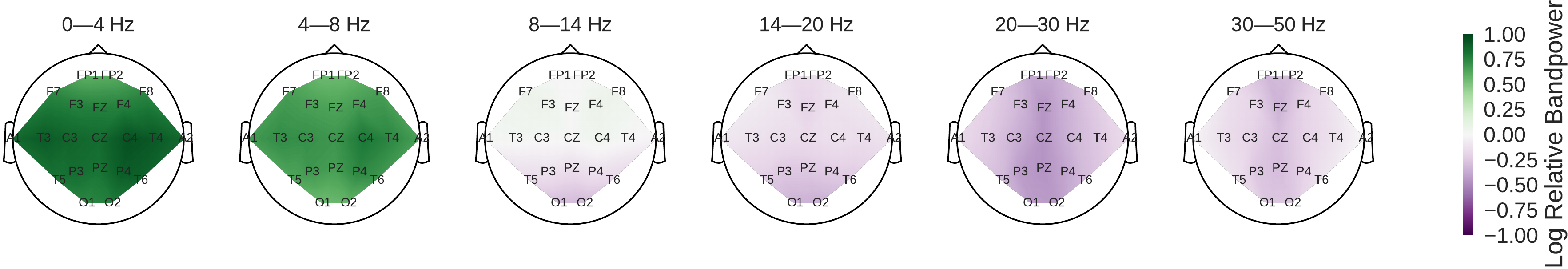}
\label{fig:spectral-bandpower}
}
\vspace{0.4cm}

\includegraphics[width=0.8\textwidth]{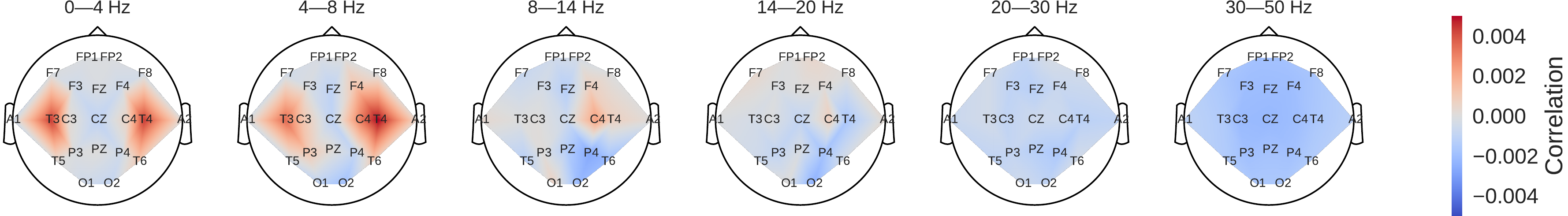}
\hfill
\subfloat[Input-perturbation network-prediction correlation maps for the deep (top) and shallow (bottom) ConvNet. Correlation of predictions for the pathological class with amplitude perturbations. Scalp maps revealed for example a bilateral positive correlation for the delta and theta frequency ranges and a spatially more broadly distributed negative correlation for the beta and low gamma frequency ranges, indicating that the ConvNets used these frequency components in their decisions
]{%
\includegraphics[width=0.8\textwidth]{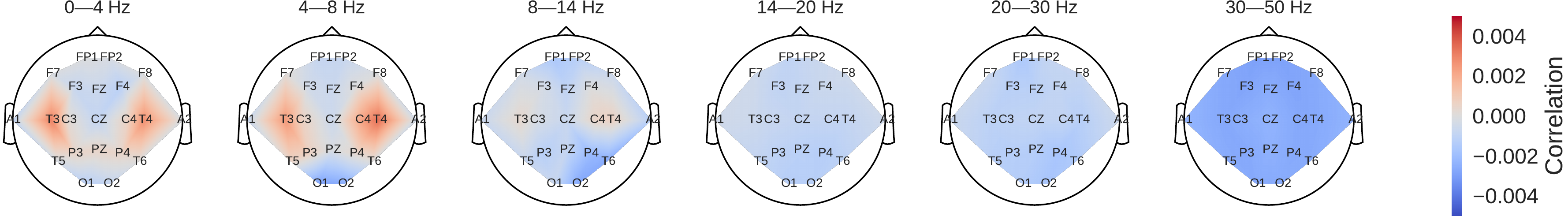}
}
\caption{\textbf{Spectral power differences and input-perturbation network-prediction correlation maps.}}
\label{fig:visualization-results}
\end{figure*}

Before moving to ConvNet visualization, we examined the spectral power changes of pathological compared to normal recordings.
Power was broadly increased for the the pathological class in the low frequency bands (delta and theta range) and decreased in the beta and low gamma ranges (Fig. \ref{fig:spectral-bandpower}).
Alpha power was decreased for the occipital electrodes and increased for more frontal electrodes.

Scalp maps of the input-perturbation effects on predictions for the pathological class for the different frequency bands showed effects consistent with the power spectra in Fig. \ref{fig:spectral-bandpower}.
Both networks strongly relied on the lower frequencies in the delta and theta frequency range for their decoding decisions. 

\subsection{Insights from the textual reports of the clinicians}

Most notably, ``small'' and ``amount'' had a much larger word frequency (15.5 times larger) in the incorrectly predicted pathological recordings compared with the correctly predicted pathological recordings.
Closer inspection showed this is very sensible, as ``small amount'' was often used to describe more subtle EEG abnormalities (``small amount of temporal slowing'', ``Small amount of excess theta'', ``Small amount of background disorganization'', ``A small amount of rhythmic, frontal slowing''), as this subtlety of changes was likely the cause of the classification errors.

Secondly, other words with a notably different frequency were ``age'' (9.7 times larger) and ``sleep'' (3 occurrences in 630 words of texts of incorrectly predicted recordings, not present in texts of correctly predicted recordings).
Both typically indicate the clinician used the age of the subject or the fact that they were (partially) asleep during the recording to interpret the EEG (``Somewhat disorganized pattern for age'', ``Greater than anticipated disorganization for age.'', ``A single generalized discharge noted in stage II sleep.''). 
Obviously, our ConvNets trained only on EEG do not have access to this context information, leaving them at a disadvantage compared to the clinicians and highlighting the potential of including contextual cues such as age or vigilance in the training/decoding approach.

Inspection of the textual records of misclassified normal recordings did not provide much insight, as they are typically very short (e.g., ``Normal EEG.'', ``Normal EEG in wakefulness.'').

Finally, consistent with the strong usage of the delta and theta frequency range by the ConvNets as seen in the input-perturbation network-prediction correlation maps (Fig. \ref{fig:visualization-results}), ``slowing'' and ``temporal'' are the 6th and 10th most frequently occurring words in the textual reports of the pathological recordings, while never occurring in the textual reports of the normal recordings (irrespective of correct or incorrect predictions).
\section{Discussion}

To the best of our knowledge, the ConvNet architectures used in this study achieved the best accuracies published so far on the TUH EEG Abnormal Corpus.
The architectures used were only very slightly modified versions of ConvNet architectures that we previously introduced to decode task-related information. This suggests that these architectures might be broadly applicable both for physiological and clinical EEG.
The identification of all-round architectures would greatly simplify the application of deep learning to EEG decoding problems and expand their potential use cases. 

Remarkably, the ConvNets already reached good accuracies based on very limited time segments of the EEG recordings. 
Further accuracy improvements could thus be possible with improved decoding models that can extract and integrate additional information from longer timescales.
The exact nature of such models, as well as the amount of EEG they would require, remains to be determined.
More accurate decoding models could either be ConvNets that are designed to intelligently use a larger input length or recurrent neural networks, since these are known to inherently work well for data with information both on shorter and longer term scales.
Furthermore, combinations between both approaches, for example using a recurrent neural network on top of a ConvNet, as they have been used in other domains like speech recognition \cite{li_constructing_2015, sainath_convolutional_2015,sak_fast_2015}, are promising.

Our automated architecture optimization provided interesting insights by yielding configurations that were markedly different from our hand-engineered architectures, yet reached similar accuracies. Since the marked improvements in training performance did not improve the evaluation accuracies in this study, in future work, we plan to use more training recordings in the optimization and study different cross-validation methods to also improve evaluation accuracies.
A full-blown architecture search \cite{mendoza_towards_2016, miikkulainen_evolving_2017, real_large-scale_2017, zoph_neural_2016, zoph_learning_2017} could also further improve accuracy. With such improved methods it would also be important not only to decode pathological vs. normal EEG in a binary fashion, but to also evaluate the possibility to derive more fine-grained clinical information, such as the type of pathological change (slowing, asymmetry, etc) or the likely underlying disorder (such as epilepsy).  

Any of these or other improvements might eventually bring the machine-learning decoding performance of pathological EEG closer to human-level performance.
Since clinicians make their judgments from patterns they see in the EEG and other available context information, there is no clear reason why machine learning models with access to the same information could not reach human-level accuracy.
This human-level performance is a benchmark for decoding accuracies that does not exist for other brain-signal decoding tasks, e.g. in decoding task-related information for brain-computer interfaces, where there is inherent uncertainty what information is even present in the EEG and no human-level benchmark exists.

Our perturbation visualizations of the ConvNets' decoding behavior showed that they used spectral power changes in the delta (0-4 Hz) and theta (4-8 Hz) frequency range, particularly from temporal EEG channels, possibly alongside other features (Fig. \ref{fig:visualization-results}). This observation is consistent both with the expectations implied by the spectral analysis of the EEG data (Fig. \ref{fig:spectral-bandpower}) and by the textual reports that frequently mentioned ``temporal'' and ``slowing'' with respect to the pathological samples, but never in the normal ones.
Our  perturbation visualization showed results that were consistent with expectations that the ConvNets would use the bandpower differences between the classes that were already visible in the spectra to perform their decoding.
Similarly, the textual reports also yielded plausible insights, e.g., that ``small amounts'' of abnormalities as indicated in the written clinical reports were more difficult for the networks to decode correctly.
Additionally, inspection of the textual reports also emphasized the importance of  integrating contextual information such as the age of the subject.

Still, to yield more clinically useful insights and diagnosis explanations, further improvements in ConvNet visualizations are needed.
Deep learning models that use an attention mechanism might be more interpretable, since these models can highlight which parts of the recording were most important for the decoding decision.
Other deep learning visualization methods like recent saliency map methods \cite{kindermans_patternnet_2017, montavon_methods_2017} to explain individual decisions or conditional generative adversarial networks  \cite{mirza_conditional_2014, springenberg_unsupervised_2015} to understand what makes a recording pathological or normal might further improve the clinical benefit of deep learning methods that decode pathological EEG.

\subsection*{Conclusion}
In summary, the deep ConvNets as presented in this study yielded the best accuracies published so far on the largest available dataset for decoding EEG pathology and by that, made a next step towards clinically useful automated EEG diagnosis.

\clearpage

\bibliographystyle{IEEEtran}
\bibliography{IEEEabrv,bibliography}

\end{document}